\documentclass[accepted]{uai2026} 
                        
\usepackage[american]{babel}

\usepackage{natbib} 
\usepackage{mathtools} 
\usepackage{amssymb}
\usepackage{booktabs} 
\usepackage{tikz} 

\usepackage{booktabs}    
\usepackage{siunitx}     
\usepackage{xcolor}      

\let\origc\c
\DeclareRobustCommand\c{\ifmmode\mathbf{c}\else\expandafter\origc\fi}
\let\origd\d
\DeclareRobustCommand\d{\ifmmode\mathbf{d}\else\expandafter\origd\fi}
\let\origu\u
\DeclareRobustCommand\u{\ifmmode\mathbf{u}\else\expandafter\origu\fi}
\let\origv\v
\DeclareRobustCommand\v{\ifmmode\mathbf{v}\else\expandafter\origv\fi}

\providecommand{\cD}{\mathcal{D}}

\providecommand{\cL}{\mathcal{L}}

\providecommand{\cN}{\mathcal{N}}

\providecommand{\cY}{\mathcal{Y}}

\providecommand{\be}{\boldsymbol{\eta}}

\providecommand{\Ex}{\mathbb{E}}

\providecommand{\KL}{\textrm{KL}}

\providecommand{\norm}[1]{\lVert#1\rVert}

\def\be{\begin{equation*}}
        \def\ee{\end{equation*}}

\title{No Data Wasted: A Semi-supervised Generative Model for Incomplete Multi-view Data Integration with Missing Labels}

\author[1]{\href{mailto:<yiyang-shen@uiowa.edu>}{Yiyang Shen}{}}
\author[1]{\href{mailto:<weiran-wang@uiowa.edu>}{Weiran Wang}{}}
\affil[1]{%
    Department of Computer Science\\
    University of Iowa\\
    Iowa City, Iowa, USA
}
  
\begin{document}
\maketitle

\begin{abstract}
  Multi-view learning is widely applied to real-life datasets, but it often suffers from both missing views and missing labels. Prior probabilistic approaches addressed the missing view problem by using a product-of-experts scheme to aggregate representations from present views and achieved superior performance over deterministic classifiers, using the information bottleneck (IB) principle. However, the IB framework is inherently fully supervised and cannot leverage unlabeled data. In this work, we propose a semi-supervised generative model that utilizes both labeled and unlabeled samples in a unified framework. Our method maximizes the likelihood of unlabeled samples to learn a latent space shared with the IB on labeled data. We also include modality-specific information in likelihood modeling and perform cross-view mutual information maximization in the shared latent space to enhance the extraction of shared information across views. Compared to existing approaches, our model achieves better predictive and generation performance on complex datasets with missing views and limited labeled samples.
\end{abstract}
\section{Introduction}
\label{sec:intro}
\vspace*{-2ex}

Many machine learning (ML) tasks involve data from multiple modalities or ``views" (used interchangeably). A sample can have multiple views or modalities to describe the same underlying identity. For example, a cartoon illustration of a cat and a photo of a cat are of different modalities but constitute the same latent characteristics of a cat and label (a cat, instead of a dog)~\citep{sangkloy2016sketchy,girella2025sketchy}. In molecular biology, DNA methylation and mRNA expression values can describe the same underlying physiological condition of the subject and predict their outcome, such as disease type and mortality. Other highly relevant application scenarios include joint representations of text, video, and audio in affective computing~\citep{zadeh2016mosi,zadeh2018multimodal,castro2019towards,hasan2019urfunny} and video, audio, and optical flow in human action video data~\citep{kay2017kinetics}.
The multi-view framework is useful even in samples with a single view. For example, one may use semantically similar parts or different augmentations of the sample as two views to extract robust representations for downstream tasks~\citep{velickovic2018deep,chen2020simple,caron2020unsupervised,tian2020contrastive,bardes2022vicreg, tian2020contrastive,tong2024empssl}.

However, real-world multi-view datasets often contain missing views, and the missing patterns can be arbitrary. 
For example, multi-omics data are multi-modal by nature and can be used to understand the underlying molecular disease mechanisms~\citep{subramanian2020multiomics}, but due to limited experimental resources, clinical designs, and sequencing platform differences, these types of data often contain arbitrary missing views and/or missing labels, and discarding such samples reduces sample sizes, waste resources, and can make downstream analysis less reliable~\citep{ron2019nemo,zhao2023clclsacrossomicslinkedembedding}. This motivates the need for a unified probabilistic models capable of learning robust shared representations, handling missing views, and supporting both prediction and cross-view generation.

Latent variable models are able to learn robust 
representations from multiple modalities with missing data, and enable prediction and generation. A popular option is to use product-of-experts (PoE) to aggregate the learned posteriors because it handles missing views and the distributions are sharper, leading to capabilities for specialized tasks \citep{hinton2002poe,wu2018multimodal,lee2021variational}. However, PoE fusion does not explicitly regularize shared information, causing information irrelevant to the task to be included, corrupting the learned representation's quality. Here, we propose adding a cross-view mutual information regularization to a semi-supervised PoE framework with private variables to address the partial multi-view representation learning problem. 

\paragraph{Our Contributions}
\begin{enumerate} 
    \item We propose a unified, PoE-based semi-supervised framework for integrated multi-view prediction and generation. 
    Our approach leads to better prediction performance than fully supervised methods, which cannot utilize unlabeled data to enforce the hidden structure relevant to prediction. It is also more convenient than an unsupervised setup, which requires representation learning followed by classifier training in two stages, potentially leading to a loss of predictive information in the first stage. 
    \item We introduce a cross-view mutual information (MI) regularization with a cycle consistency-style imputation that efficaciously learns a robust joint latent representation in the missing view setup. 
    Through an empirical comparison to prior state-of-the-art posterior view-agreement regularization methods, we demonstrate that cross-view MI regularization consistently achieves superior performance across diverse datasets.
    \item By introducing private variables and deriving a new ELBO that facilitates better generation, we show that the jointly learned representations in our framework enhance the quality of learned representations, evidenced by improved prediction and conditional imputation.
\end{enumerate}


\section{Methods}
\label{sec:methods}
\vspace*{-1ex}


\paragraph{Notations}
Our dataset consists of a total of $V$ input views with set $\mathcal{V}$. 
Let $x_v$ be the $d_v$-dimensional input feature from the $v$-th view and $y \in \cY$ be the output label. 
We denote $x_v=\varnothing$ if the $v$-th view is missing, $y=\varnothing$ if the label is missing, and input features collectively as $X=\{x_v: v\in \mathcal{V}\}$.
The set of \emph{observed views} for $X$ is 
denoted by $\mathcal{S}=\{v:x_v\neq \varnothing\}\subseteq \mathcal{V}$, with the associated input features being $X_{\mathcal{S}}=\{x_v: v\in \mathcal{S}\}$. A data sample $(X, y)$ is incomplete if $x_v=\varnothing$ for some view $v$ and/or $y=\varnothing$.

We have a dataset $\cD=\{X_n, y_n\}_{n=1}^N$ where each sample $(X_n, y_n)$ may be incomplete but at least one view is observed. The set of samples with observed labels are referred to as the supervised portion whereas the rest unsupervised. 

\subsection{Problem Statement}
\label{sec:setup}
\vspace*{-1ex}

We assume different views of the same sample offer the same predictive information (e.g., cat illustration and cat photo) and term it the ``multi-view redundancy assumption''~\citep{federici2020learning}. With this assumption in mind, our goal is to extract a robust latent representation $z$ from a multi-view observation $X$ with an arbitrary view-missing pattern to predict target $y$. We focus on the semi-supervised setup where ground truth labels are only present for a small portion of the dataset, so we need to leverage a large amount of unlabeled data to learn the discriminative information. We use variational methods to extract a common subspace shared by the views, based on which we perform both reconstruction of missing inputs and prediction of discrete labels. Our setup is a generalization of that of DeepIMV~\citep{lee2021variational} by allowing missing labels. 

The incomplete multi-view problem involves two major challenges. 
First, the learned representations must integrate incomplete observations with various view-missing and label-missing patterns in a unified framework.
Second, the learned representation must integrate information shared across views \emph{that are relevant to the prediction task}. 
In the rest of this section, we provide the components of our model architecture and learning objective, which pave the way for tackling the above challenges.

\subsection{PoE-based Joint View Representation}
\vspace*{-1ex}

To obtain joint representation of a set of available views $\mathcal{S} \subseteq \mathcal{V}$ with arbitrary missing pattern, one can realize a common latent space by probabilistically encoding each view with the posterior distribution $q_{\theta}(z_v|x_v)$~\citep{kingma2019introduction} and aggregating all encoded present views into a joint representation $q_\theta(z|X)$.
One particular elegant aggregation scheme is product-of-expert (PoE), defined as follows~\citep{wu2018multimodal}:
\begin{equation}\label{eq:poe}
    \begin{aligned}
        q_\theta(z|X_{\mathcal{S}}) 
        \propto p(z)\prod_{v\in \mathcal{S}} q_{\theta}(z|x_v).
    \end{aligned}
\end{equation}
When $q_{\theta}(z|x_v)$ are parameterized as Gaussian distributions, $q_\theta(z|X_{\mathcal{S}})$ is also Gaussian, whose mean and variance can be computed in closed-form from local posteriors of the present views. 
PoE is well-suited for our setup because it does not require missing-view imputation, and has been shown to provide good generation quality~\citep{daunhawer2022limitation}. 

More recently, \citet{sutter2021generalized} proposed a generalization of both PoE and mixture-of-experts (MoE, defined as $q_\theta(z|X) = \frac{1}{V}\sum_{v=1}^V q_\theta(z_v|x_v)$), named mixture-of-product-of-experts (MoPoE): 
\begin{gather}\label{eq:mopoe}
q_\theta(z|X_{\mathcal{S}})
    \propto \sum_{\mathcal{O}\subseteq \mathcal{S}} \; \prod_{v\in \mathcal{O}} q_{\theta}(z|x_v),
\end{gather}
where we enumerate all subsets of $\mathcal{S}$.
An advantage of MoPoE is that, by further taking subsets of $S$ in the posterior definition and evidence lower bound (ELBO), we simulate more (and severer) missing pattern for each training sample and therefore potentially improve the generation. As we will see in Sec~\ref{sec:mnist_crossgen}, this also allows us to incorporate prior in likelihood modeling and facilitates cross-view generation.
During training, to alleviate the cost of computing the full sum in~\eqref{eq:mopoe}, we randomly sample a single subset of $\mathcal{S}$ when estimating the loss for every minibatch.

\subsection{Loss Components}
\vspace*{-1ex}

To address the challenges mentioned in Sec~\ref{sec:setup}, we propose a semi-supervised representation learning framework that consists of the following loss components. 
The overall architecture is shown in Figure~\ref{fig:architecture}.

\subsubsection{Supervised Information Bottleneck}
\label{sec:sup_ib}
\vspace*{-1ex}

In the supervised data portion $(X, y)$, we estimate the predicted label $y$ based on $q_{\psi}(y|z)$, given the latent sample $z$ drawn from $q_\theta(z|X_{\mathcal{S}})$, where $\psi$ represents predictor parameters.

To ensure the robustness and generalization of the prediction model $X \rightarrow z \rightarrow y$, the information bottleneck (IB,~\citealt{tishby1999information}) principle maximizes task-relevant information in $z$, i.e., $I(y;z)$ to correctly predict the label $y$ and minimize redundant, or irrelevant, information $I(X, z)$ to reduce the effect of nuisance factors in data.
Because computing $I(y;z)$ requires an explicit target, this framework is fully supervised. One can derive the following variational objective based on the IB principle \citep{lee2021variational}:
\begin{align}\notag
    \cL_{\text{joint}}^{\theta, \psi}(X,y)
    =& - \alpha\cdot I(y;z) + \gamma \cdot I(X;z)\\
    \notag
    \approx&\; \alpha \cdot \Ex_{z\sim q_{\theta}(z|X_{\mathcal{S}})}\bigl[-\log q_\psi (y|z)\bigr] \\\label{eq:sup_joint_loss}
    &+\gamma \cdot \KL [q_\theta(z|X_{\mathcal{S}})\|p(z)],
\end{align}
where $\theta$ refers to the inference parameters in encoders which provide the approximated posteriors, the first term is the cross-entropy loss between prediction and ground truth $y$ with classification parameters $\psi$, and the second term is the Kullback-Leibler (KL) divergence between the joint posterior and prior $p(z)$, a distribution independent from the data, e.g., a standard isotropic Gaussian. The loss for every observed view $v\in \mathcal{S}$ can be derived in a similar fashion:
\begin{equation}\label{eq:sup_marg_loss}
\begin{aligned}
    \cL_{\text{marginal}}^{\theta, \psi}(x_v,y) =& - \beta \cdot I(y;z_v) + \gamma_v \cdot I(x_v;z_v)\\
    \approx &\;\beta\cdot \Ex_{z_v\sim q_{\theta}(z_v|x_v)}\bigl[-\log q_\psi (y|z_v)\bigr] \\
    & + \gamma_v \cdot  \KL [q_\theta(z_v|x_v)\|p(z_v)],
\end{aligned}
\end{equation}
where empirically we set $\gamma_v=\gamma$. Essentially, we have a joint classifier on $z$ and $|\mathcal{S}|$ marginal classifiers on $z_v$'s. Combining~\eqref{eq:sup_joint_loss} and~\eqref{eq:sup_marg_loss}, the supervised loss becomes:
\begin{equation}\label{eq:sup_loss}
    \cL_{\text{sup}}^{\theta, \psi}(X,y)=\cL_{\text{joint}}^{\theta, \psi}(X,y)+\sum_{v\in \mathcal{S}} \cL_{\text{marginal}}^{\theta, \psi}(x_v,y).
\end{equation}

\subsubsection{Likelihood Modeling with Missing Views}
\vspace*{-1ex}

With limited number of labels, the fully supervised IB loss may fail to extract a robust representation. A straightforward solution is to extract a representation based on unsupervised data, hoping that the representation captures information on unobserved labels~\citep{bengio2013representation}. \citet{wu2018multimodal} proposed an ELBO for observed data, using the aggregated posterior of available views, i.e.,
\begin{align*}\nonumber
    \text{ELBO}^{\theta,\phi}(X) =& \Ex_{z \sim q_\theta (z|X_{\mathcal{S}})}\Bigl[\sum_{v\in \mathcal{S}} \log p_\phi (x_v|z)\Bigr]  \\
    & + \gamma\cdot \KL\bigl[q_{\theta}(z|X_{\mathcal{S}})\|p(z)\bigr],
\end{align*}
where $\phi$ denotes the likelihood parameters in decoders.
However, the above ELBO only depends on the shared variable $z$ and cannot recover information that is specific to the target modality~\citep{daunhawer2022limitation}.
Therefore, inspired by previous work on multi-view representation learning~\citep{wang2016vcca,daunhawer2020self,lee2021privateshared,zhang2026idmvae}, we propose to use 
both a shared latent variable $z$ and modality-specific (private) latent variables $w_v$, $v=1,\dots,V$.
Ideally, we would like to achieve a dichotomy of two types of variations: $z$ should capture shared variations across available views, while $w_v$ should capture information specific to modality $v$. This way, we can hope to improve task prediction performance by removing nuisance factors from $z$ and enable high quality, controllable generation (e.g., we can swap either shared or private variable of an sample to generate a sample sharing desired variations).

While $w_v$ can capture potentially rich private variations of $x_v$, it also leads to a challenge for cross-view generation/imputation:  
although we could extract the shared information from present views, we have no access to the posterior $q(w_v|x_v)$ when $x_v$ is missing, and must use the prior $p(w_v)$ for filling in the private information. 
Thus, to obtain high quality generation, we must involve the prior during likelihood modeling, beyond KL regularization.

\subsubsection{A New ELBO for MoPoE with Missing Views}
\label{sec:new_elbo_mopoe}
\vspace*{-1ex}

To address the above challenge, we propose a new ELBO  for the missing view problem in the MoPoE framework.
Let $\mathcal{S}$ be the set of all available views, and $\mathcal{O} \subset \mathcal{S}$ is a true subset we use in~\eqref{eq:mopoe} 
for estimating $q(z|X)$.
We approximate the expected log-likelihood $\Ex_{q(z|X_{\mathcal{O}})} [ \log p_\phi(x_v|z) ]$ differently depending on whether $v \in \mathcal{O}$.

In the first case where $v\in\mathcal{O}$, and one can lower bound the likelihood as follows:
\begin{equation}\label{eq:unsup_present}
\begin{aligned}
\log p_\phi(x_v|z)\ge &\; \Ex_{w_v\sim q_\theta(w_v|x_v)} \left[ \log {p_\phi(x_v|z,w_v)} \right] \\
    &- \KL[q_\theta(w_v|x_v)||p(w_v)].
\end{aligned}
\end{equation}
In other words, we use the posterior of $w_v$ in reconstruction.

In the second case where $v\in \mathcal{S}\setminus\mathcal{O}$, one can still train the likelihood model for $x_v$ using the posterior $q(z|X_{\mathcal{O}})$ and the prior $p(w)$ for private information, i.e.,
\begin{equation}\label{eq:unsup_missing}
\begin{aligned}
  \log  p(x_v|z)\ge \Ex_{p(w_v)} [\log  p(x_v|w_v,z)].
\end{aligned}
\end{equation}
Combining the above bounds in~\eqref{eq:unsup_present} and~\eqref{eq:unsup_missing}, we obtain a new ELBO loss (full derivation in Appendix \ref{apdx:elbo}):
\begin{gather}
\label{eq:unsup_loss}
\cL_{\text{unsup}}^{\theta,\phi}(X) = \notag\\  
-\eta \cdot \Ex_{z\sim q_\theta(z|X_{\mathcal{O}})}
 \sum_{v\in \mathcal{O}}
   \Bigl\{
      \Ex_{w_v\sim q_\theta(w_v|x_v)}
      [\log p_\phi(x_v|z,w_v)] \notag\\
 + \frac{\gamma}{\eta} \cdot \KL[q_\theta(w_v|x_v)\|p(w_v)]
   \Bigl\} \notag\\
   - \eta \cdot \Ex_{z\sim q_\theta(z|X_{\mathcal{O}})}
   \sum_{v \in \mathcal{S} \setminus \mathcal{O}}
      \Bigl\{\Ex_{w_v\sim p(w_v)}
      [\log p_\phi(x_v|z,w_v)] \Bigl\}
      \notag\\
 + \gamma \cdot \KL[q_\theta(z|X_{\mathcal{O}})||p(z)],
\end{gather}
where KL terms has a coefficient of $\gamma$ with the same magnitude defined in the supervised loss.

While one can use simple Gaussian for $p(w_v)$, for structured latent space (e.g., with clustering structure), we use a diffusion process prior~\citep{ho2020denoising,vahdat2021score} to increase the capacity of prior and obtain high quality generations (Sec~\ref{sec:mnist_crossgen}).

\subsubsection{Cross-view Mutual Information Maximization}
\label{sec:cvmi}

In Sec \ref{sec:sup_ib}, one can maximize $I(y;z)$ so that the learned representation contains predictive information. However, when the label $y$ is not observed, one can guarantee that $z_1$ is sufficient for predicting $y$ without knowing $y$, if two views provide the same predictive information (multi-view redundancy assumption), i.e., $I(y;x_1|x_2)=I(y;x_2|x_1)=0$.
Therefore, any representation containing all the information shared by both views is as predictive as their joint observation~\citep{federici2020learning}. 

This motivates us to use $x_2$ as a proxy of $y$. With this variable substitution, we can instead enforce $z_1$ to contain high mutual information with $x_2$. This MI can be lower bounded as $I(z_1;x_2)\geq I(z_1;z_2)$,
where $z_2$ is the representation of $x_2$ modeled by $q(z_2|x_2)$.
We can therefore extract shared information by maximizing $I(z_1;z_2)$.
For a detailed discussion, see Appendix \ref{apdx:cvmi}.

To estimate $I(z_1;z_2)$, we use the contrastive estimate of mutual information~\citep{oord2018representation}:
\begin{align} \label{eqn:contrastive-mi}
I(z_1;z_2) \approx 
\Ex_{z_1,z_2} \log \left[\frac{s(z_1,z_2)}{s(z_1,z_2) + \sum_{j=1}^n s(z_1,\bar{z}_2^j)}\right],
\end{align}
where $s(z_1, z_2) = \exp \left( \frac{z_1^\top z_2}{\norm{z_1} \cdot \norm{z_2}} \right)$ is an affinity function, and $\bar{z}_2^j$ are $n$ randomly sampled negative examples from the same minibatch not aligned with $z_1$.
We ensure a large enough batch size to obtain a good estimate of~\eqref{eqn:contrastive-mi}. It is possible to introduce other hyperparameters (e.g., temperature in affinity function) and use hard negative examples~\citep{robinson2021hardneg,jiang2025negsampl} to further improve performance, but we do not pursue them here.
In~\eqref{eqn:contrastive-mi}, the expectation is computed over samples when both view 1 and 2 are observed. 
In practice, we use the symmetric version of~\eqref{eqn:contrastive-mi} where we also switch the role of the two views and average these two estimates. Intuitively, this enforces the latent representation for paired data $(x_1, x_2)$ to be similar while those of unpaired data are pushed away.

When our dataset consists of more than two views, we can maximize the average of the above contrastive estimate for all pairs of views~\citep{tian2020contrastive,zhang2026idmvae}. 
We can therefore minimize the following term 
\begin{equation}\label{eq:cvmi}
\cL_{\text{cvmi}}(z)=- \zeta \cdot \frac{1}{\binom{V}{2}}\sum_{i\neq j} I(z_i;z_j),
\end{equation}
where $\zeta$ is the regularization coefficient.

For many multi-view learning problems, the multi-view redundancy assumption is approximately satisfied as evidenced by our experimental results, where $\cL_{\text{cvmi}}(z)$ can significantly boost the performance when likelihood modeling alone is challenging. 

\begin{figure}[t]
\centering
\includegraphics[trim={2.2cm 8.8cm 6.0cm 3.5cm}, clip=true, width=\columnwidth]{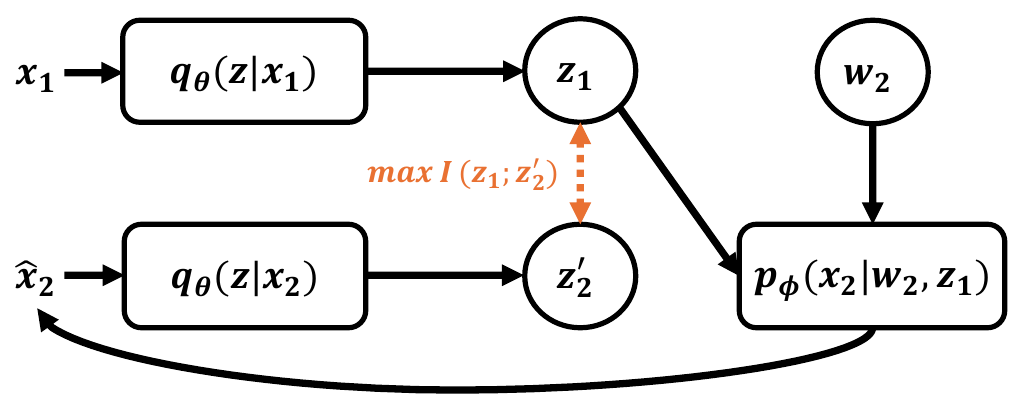}
\caption{Our scheme to impute representation $z'_2$ when view 2 is missing for a sample, in order to compute $\cL_{\text{cvmi}}$.
}
\label{fig:architecture-imp}
\end{figure}

\paragraph{Imputed Representations for Contrastive Estimators.}
\label{sec:imp_mi}

We now extend the above regularization to the missing view problem to create additional training signals for the model.
When one view $z_i$ of the representation pair $(z_i,z_j)$ is unavailable, we propose a cycle consistency-style heuristic for imputation $z_i'\sim q(z|f_{\phi}^i (w_i^\prime, z_j))$. Here the latent sample $w_i^\prime$ is either an uninformative prior sample or the private variable extracted from another sample with \emph{present} view $i$, and $f_{\phi}$ denotes the decoder of view $i$,
with which we can synthesize a new sample $f_{\phi}^i (w_i^\prime, z_j))$. While we do not have the ground truth observation of this missing view, when we map it back to the latent space, the shared variable should ideally remain unchanged.
In this way, we can compute $I(z_i',z_j)$ even with one missing representation in the pair, so \eqref{eq:cvmi} can be computed as long as at least one of the views in the pair is available.
See Figure~\ref{fig:architecture-imp} for a graphical explanation.
As shown in Appendix \ref{apdx:ablation}, this imputation helps boost latent classification performance. 

\paragraph{Previous View-agreement Regularization.} 
Other regularization terms for promoting view-agreement were proposed in the multi-view representation learning literature. 

\citet{hwang2021tc} proposed a conditional variational IB (CVIB) regularization:
\begin{equation*}
\cL_{\text{cvib}} (z) = \sum_{v\in\mathcal{S}} \KL[q(z|X_{\mathcal{S}})||q(z_v|x_v)],
\end{equation*}
which tries to address the issue of unbalanced representation where the joint posterior relies on informative unimodal encoders and ignores the rest, by incentivizing agreement between the PoE aggregation and 
view-specific encoders. 

Inspired by VAMP-prior~\citep{tom2018vampprior},~\citet{sutter2024unity} proposed to replace the simple prior $p(z)$ with an MoE ``prior'' $\frac{1}{M}\sum_{v} q(z_v|x_v)$ for the generative model. As a result, the KL term of the ELBO becomes: 
\begin{equation*}
    \cL_{\text{mmvm}} (z) = \sum_{v\in \mathcal{S}} \KL\Bigl[q(z_v|x_v) \big\| \frac{1}{|\mathcal{S}|}\sum_{v'\in \mathcal{S}} q(z_v'|x_v')\Bigr].
\end{equation*}
We refer to this regularization as MMVM.

These two methods share the same intuition, in that the per-view encoder should be aligned with the joint posterior. Note that, these works do not explicitly model view-specific information like we do.
A key contribution of this work is a systematic empirical evaluation of these view-agreement regularization methods.


\begin{figure}[t]
\centering
\includegraphics[trim={1.8cm 1.8cm 2.0cm 2.0cm}, clip=true, width=\columnwidth]{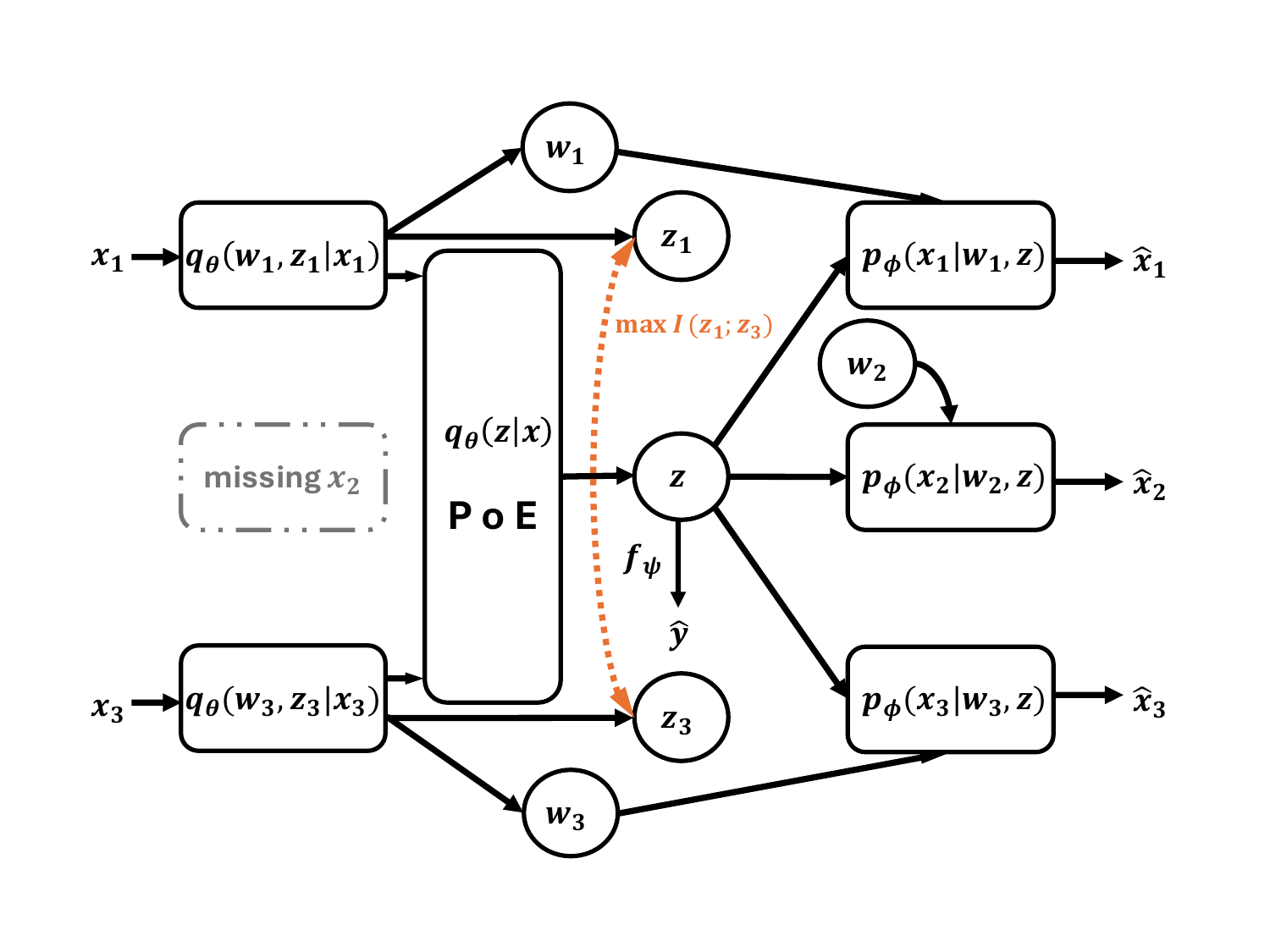}
\caption{The architecture of our method. $x_2$ is missing and $w_2$ is an uninformative prior sample concatenated with $z$ for generation of the missing view $x_2$. View-specific classifiers are omitted for brevity. Other variables are explained throughout Sec \ref{sec:methods}.
}
\label{fig:architecture}
\end{figure}

\subsubsection{Overall Objective} Our final objective combines the above intuitions:
\begin{equation}
\arg\min_{\phi,\psi,\theta} \cL_{\text{total}}(X,y)
=\cL_{\text{sup}}^{\theta, \psi}(X,y)+\cL_{\text{unsup}}^{\theta,\phi}(X)+ \cL_{\text{cvmi}}^{\theta,\phi}(z)
\end{equation}
An illustration of our model is given in Figure \ref{fig:architecture}.


\section{Related Work}
\label{sec:related}
\vspace*{-2ex}

\textbf{Representation Learning with Missing Views.}
Current missing modalities handling techniques can be categorized into generative methods and methods based on latent space alignment (or view-agreement regularization in our terminology)~\citep{nguyen2025robult}. 

For the former, unsupervised methods typically used to learn multi-view representations in a missing-view setup were proposed using different posterior aggregation schemes, such as mixture of experts (MoE)~\citep{shi2019variational}, PoE~\citep{wu2018multimodal}, mixture-of-product-of-experts (MoPoE)~\citep{sutter2021generalized}, CPM-Nets~\citep{zhang2019cpmnets}, sum- and attention- pooling~\citep{hirt2024perminv}. However, these methods by themselves require 
two training stages--representation learning followed by supervised training, 
making them less convenient, and it is possible that predictive information is lost during the first stage.

For the latter, \citet{hwang2021tc} and \citet{sutter2024unity} propose unsupervised methods to align unimodal posterior to aggregated posteriors, but enforcing such agreement does not necessarily capture useful information, especially when such information is sparse. 

Because of these limitations, semi-supervised methods are especially appealing as they can perform prediction and imputation in a unifying manner, while potentially addressing the missing view problem. Relevant Graph Neural Network-based methods for missing-view, missing-label data such as MUSE~\citep{wu2024muse} and DIMvLN~\citep{jiang2024mvnetwork} were proposed, but they require dataset-dependent graph constructions and therefore domain experts (e.g., healthcare professionals) for setup, making it less applicable in a general setting. 


\textbf{Information-Theoretic Representation Learning.}
The IB principle~\citep{tishby1999information} provides a theoretical framework for learning compact and informative representations of data. 
It proposes a variational IB method to find a representation $z$ of the input data $X$ that maximizes the mutual information (MI) between $z$ and the target variable $Y$ for prediction, while minimizing the mutual information between $z$ and $X$ which could be nuisance factors.
\citet{lee2021variational} applies this method to the missing view problem in a fully supervised setup and obtained impressive performance. \citet{federici2020learning} exploited the multi-view redundancy assumption to leverage one of the two learned views as label, rendering unsupervised learning with the goal of prediction possible. Following this path, \citet{wang2025information} proposed a deterministic two-stage learning scheme to first learn a shared representation in the first stage before modality-specific information in the second stage with theoretical guarantees. Most recently, \citet{zhang2026idmvae} used shared latent space alignment proposed in \citet{federici2020learning} along with an MI augmentation scheme to achieve information-theoretic empirical disentanglement between shared and private variables.
\section{Experiments}
\vspace{-1ex}

\begin{table*}[t]
  \centering
  \caption{%
    Prediction performance of TCGA with 5\% (left) and 1\% (right) available labels.
  }
  \label{tab:pred_tcga}
  \vspace*{-1ex}
  \begin{minipage}[t]{0.48\textwidth}
    \vspace{0pt}
    \centering
    \begin{tabular}{@{}l c c@{}}
    \hline
    \textbf{Method}    & \textbf{Accuracy}  & \textbf{AUROC}   \\
    \hline
    \textit{Supervised Methods}\\
    Sup Base                         &   0.690$\pm$0.042       &   0.711$\pm$0.041        \\
    DeepIMV             &     0.647$\pm$0.017            &   0.754$\pm$0.024       \\
    \midrule
    \textit{Unsupervised Methods}\\
    MVAE                     &   0.641$\pm$0.016           &   0.557$\pm$0.024      \\
    DMVAE                    &   0.641$\pm$0.016            &  0.572$\pm$0.038        \\
    \midrule
    \textit{Semi-supervised Methods} \\
    Semi-sup Base +w  & 0.645$\pm$0.011    & 0.755$\pm$0.017             \\
    CVIB +w           &   0.643$\pm$0.013                  &   0.755$\pm$0.017       \\
    MMVM +w        &   0.647$\pm$0.017           &   0.750$\pm$0.024       \\
    Ours  +w & \textbf{0.717$\pm$0.025}     & \textbf{0.779$\pm$0.011}  \\
    \hspace{2em} –w       & 0.705$\pm$0.014         & 0.757$\pm$0.028  \\

    \hline
  \end{tabular}
\end{minipage}%
  \hfill
  \begin{minipage}[t]{0.48\textwidth}
    \vspace{0pt}
    \centering
      \begin{tabular}{@{}l c c@{}}
    \hline
    \textbf{Method}    & \textbf{Accuracy}   & \textbf{AUROC}  \\
    \hline
    \textit{Supervised Methods}\\
    Sup Base                      &   0.624$\pm$0.037           &   0.605$\pm$0.045       \\
    DeepIMV                   &   0.653$\pm$0.026          &   0.648$\pm$0.029       \\
    \midrule
    \textit{Unsupervised Methods}\\
    MVAE                    &   0.641$\pm$0.016            &   0.523$\pm$0.017       \\
    DMVAE                       &   0.641$\pm$0.016           &   0.560$\pm$0.037      \\
    \midrule
    \textit{Semi-supervised Methods}\\
    Semi-sup Base +w    &   0.631$\pm$0.021 &   0.674$\pm$0.020         \\
    CVIB +w   &   0.642$\pm$0.015& 0.684$\pm$0.036   \\
    MMVM +w       & 0.675$\pm$0.030& 0.677$\pm$0.032  \\
    Ours +w                 & \textbf{0.677$\pm$0.040}     & \textbf{0.718$\pm$0.052}  \\
    \hspace{2em} –w  &   0.655$\pm$0.042  &   0.678$\pm$0.046         \\
    \hline
  \end{tabular}
\end{minipage}%
  \hfill
  \begin{minipage}[t]{0.48\textwidth}
    \vspace{0pt}
    \centering
  
\end{minipage}%
\vspace*{-1ex}
\end{table*}

We compare our method against the following fully supervised, unsupervised, and semi-supervised baselines on synthetic and real-world datasets. 

\vspace{-0ex}
\paragraph{Fully supervised methods.}

\vspace{-1ex}
\begin{itemize}
    \item \textbf{Supervised Base}: trains individual classifiers for every view, and performs majority voting for final prediction. 
    
    \item \textbf{DeepIMV}~\citep{lee2021variational}: uses a PoE scheme to integrate available views' posteriors and trains both view-specific predictors and joint predictors. DeepIMV is fully supervised and consequently cannot use samples with missing labels at training time. We report the joint predictor's performance.
\end{itemize}

\vspace{-1ex}
\paragraph{Unsupervised generative methods.}

\vspace{-1ex}
\begin{itemize}
    \item \textbf{MVAE}~\citep{wu2018multimodal}: learns a representation $z$ of present views using the simple PoE aggregation, and maximizes the ELBO of present views.
    \item \textbf{DMVAE}~\citep{lee2021privateshared}: learns both a shared representation $z$ using PoE, and a view-specific representation $w_v$ for every view. DMVAE maximizes an ELBO using both posteriors of $z$ and $w_v$.
\end{itemize}

\vspace{-1ex}
\paragraph{Semi-supervised methods.}

\vspace{-1ex}
\begin{itemize}
    \item \textbf{Semi-supervised Base}: performs PoE aggregation on each sample and optionally uses view-specific variables to maximize ELBO on unsupervised data, and trains a joint predictor and view-specific predictors on supervised data at the same time. No view-agreement regularization is used.

    \item \textbf{CVIB}: semi-supervised base with CVIB regularization~\citep{hwang2021tc}.
    \item \textbf{MMVM}: semi-supervised base with MMVM regularization~\citep{sutter2024unity}.
\end{itemize}
Our method can be seen as semi-supervised base with cross-view mutual information regularization.
See Sec~\ref{sec:cvmi} for definition of the view-agreement regularizations.

We determine whether to use private variable for semi-supervised methods based on the comparison of DMVAE versus MVAE. Whenever view-specific representations in unsupervised learning methods benefit latent classification, semi-supervised baselines use them as well.

\paragraph{Evaluation} Our main performance metrics is the classification performance on the shared latent representation $z$.
To obtain performance metrics of both MVAE and DMVAE, we first learn representations of unsupervised data, and then train MLP classifiers on the aggregated posteriors $p(z|X_{\mathcal{S}})$.

\begin{table}[t]
\centering
\caption{PolyMNIST-Quadrant latent classification performance with 0.02\% available labels.
}
\label{tab:pred_mnist}
  \vspace*{-1ex}
    \begin{tabular}{@{}l c c@{}}
        \hline
        \textbf{Method}   & \textbf{Accuracy}   & \textbf{AUROC}   \\
        \hline
        \textit{Supervised Methods}\\
        Sup Base                         &   0.6422          &   0.9180     \\
        DeepIMV                      &   0.8348         &   0.9851     \\
        \midrule
        \textit{Unsupervised Methods}\\
        MVAE                         &  0.1157          &   0.5209    \\
        DMVAE                &   0.1257                   &   0.5251     \\
        \midrule
        \textit{Semi-supervised Methods}\\
        Semi-sup Base –w    & 0.8843      & 0.9934       \\
        CVIB –w  &   0.9704    &   0.9997      \\
        MMVM –w    &   0.7545      &   0.9650     \\
        Ours –w &   0.9503       &   0.9974      \\
        \hspace{2em} +w    &   \textbf{1.0000}  &   \textbf{1.0000}        \\
        \hline
    \end{tabular}
\vspace*{-1ex}
\end{table}

\subsection{The Cancer Genome Atlas (TCGA)}
\vspace{-1ex}

TCGA\footnote{\url{https://www.cancer.gov/tcga}} is a real-world multi-omics dataset that contains four modalities associated to clinical labels: DNA methylation, microRNA expression, mRNA expression, RPPA (protein expression). We use one-year mortality as our prediction task and record area under the ROC curve (AUROC) and accuracy in the latent space to compare the robustness of different methods. Using five random seeds, five 90\%/5\%/5\% train/validation/test splits are generated. We also dropped 50\% of views in the training set and restored views so that there is at least one view available per sample. Lastly, we dropped 95\% and 99\% of labels in the training set to simulate scenarios with few labels, which is likely in experimental science. We use two-layer feedforward networks as encoders and decoders. The data processing and implementation details can be found in Appendix \ref{apdx:tcga}. 

Latent classification results are shown in Table \ref{tab:pred_tcga}. We observe that when there are only a few labels, supervised methods in general suffer from overfitting, which leads to poor test performance. Additionally, unsupervised methods does not perform well because they simply aggregate all information to a compact representation, without focusing on the shared information. That is, information irrelevant to the task is also 
kept in $q(z|X_{\mathcal{S}})$, hurting the performance of unsupervised methods.
Given that DMVAE outperforms MVAE, we include private variables in likelihood modeling for semi-supervised methods.
For semi-supervised methods, we show that modality-agreement regularization methods, namely CVIB and MMVM, perform similarly to semi-supervised base methods, while our methods perform better due to inductive bias introduced by contrastive learning~\citep{Tschannen2020mutual}.

Lastly, we perform an ablation study where we remove $w_v$ from likelihood modeling which leads to worse prediction accuracy (last row of tables), likely because $w_v$ alleviates $z$ from capturing all information about $x_v$ and thus $z$ can better capture the shared information. 



\begin{figure}[t]
\caption{
PolyMNIST-Quadrant conditional generation (missing view imputation). Conditioned on views 1-4 (top 4 rows), we generate view 5. Each column correspond to the same sample. Each row share the same private variable.}
\label{fig:tpmnist_gen}
    \centering
\begin{tabular}{c}
Condition, views 1-4 \\
\includegraphics[width=\linewidth]{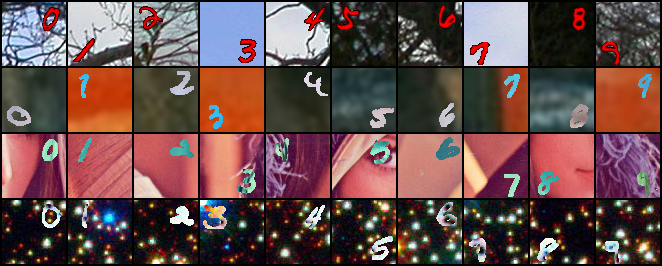} \\
Ours -w, PoE scheme \\
\includegraphics[width=\linewidth]{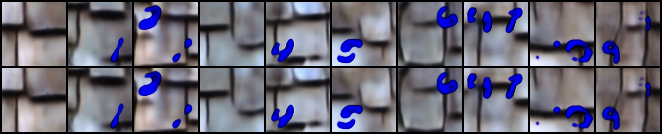} \\
Ours +w, MoPoE scheme, Simple prior for $w_v$ \\
\includegraphics[width=\linewidth]{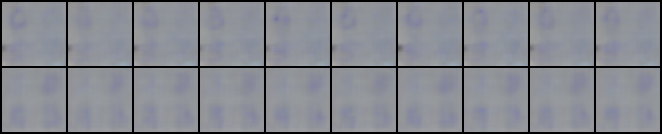} \\
Ours +w, MoPoE scheme, Diffusion prior for $w_v$ \\
\includegraphics[width=\linewidth]{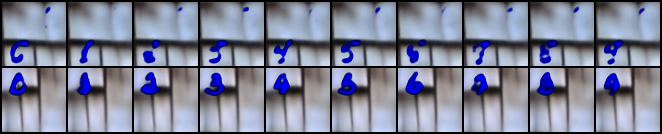} \\
CVIB +w, MoPoE scheme, Diffusion prior for $w_v$ \\
\includegraphics[width=\linewidth]{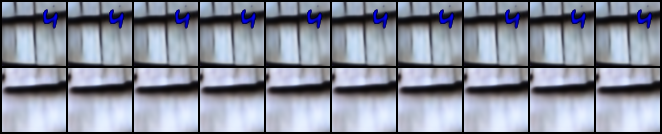}
\end{tabular}
\end{figure}

\subsection{PolyMNIST-Quadrant}
\vspace{-1ex}

PolyMNIST is a benchmarking dataset for multimodal learning~\citep{sutter2021generalized}. It consists of MNIST digits~\citep{Lecun_98a} overlaid on patches randomly cropped from five complex background photos, so there are five views. 
Here we use the more challenging variant called PolyMNIST-Quadrant~\citep{zhang2026idmvae}, which places 32x32 digits into one of the four quadrants in 64x64 canvas (i.e., the randomly cropped patch from the background). This dataset is suitable for our purpose because we want to show that both the view-specific background and shared digit identity are preserved 
when private information is dominant and shared information is hard to extract by maximum likelihood modeling.  
We drop 50\% of the views and 99.8\% of class labels. Latent classification is performed on digit identities; generation of one view is conditioned on the other four. We use ResNet~\citep{he2015deepresiduallearningimage} as encoders and decoders. See implementation details in Appendix~\ref{apdx:mnist}.

\vspace{-1ex}
\subsubsection{Latent Classification}
\vspace{-2ex}

Latent classification results in Table~\ref{tab:pred_mnist} show that DeepIMV significantly outperforms supervised base, indicating the effectiveness of PoE fusion in light of a missing view setup. This supports our motivation to improve this framework. 
For this dataset, MVAE (slightly) outperforms DMVAE, so we do not include private variables for semi-supervised methods.
Additionally, we show that, with the Cross-view MI regularization and private variables, our method was able to achieve 100\% accuracy. This is not surprising because by maximizing Cross-view MI, we learn a robust representation of digit identity that is shared across views, even when the amount of shared information is small. The other modality agreement methods, namely CVIB and MMVM, can only do so to a lesser extent.

\begin{table*}[t]
  \centering
  \caption{
    Prediction performance of UR-FUNNY (left) and CMU-MOSEI (right) with 10\% available training labels.
  }
  \label{tab:pred_multibench}
  \vspace*{-1ex}
  \begin{minipage}[t]{0.48\textwidth}
    \vspace{0pt}
    \centering
    \begin{tabular}{@{}l c c@{}}
        \hline
        \textbf{Method}     & \textbf{Accuracy}   & \textbf{AUROC} \\
        \hline
        \textit{Supervised Methods}\\
        Sup Base                     &   0.567$\pm$0.013             &   0.597$\pm$0.017      \\
        DeepIMV                 &   0.557$\pm$0.035           &   0.615$\pm$0.008        \\
        \midrule
        \textit{Unsupervised Methods}\\
        MVAE                     &  0.562$\pm$0.002           &   0.606$\pm$0.002       \\
        DMVAE              &   0.514$\pm$0.002                   &   0.580$\pm$0.008       \\
        \midrule
        \textit{Semi-supervised Methods}\\
        Semi-sup Base –w     & 0.596$\pm$0.008  & 0.617$\pm$0.009          \\
        CVIB –w  &   0.601$\pm$0.005  &   0.618$\pm$0.008        \\
        MMVM –w      &   0.591$\pm$0.015        &   0.625$\pm$0.005      \\
        Ours –w    &   \textbf{0.603$\pm$0.010} &   0.624$\pm$0.006         \\
        \hspace{2em} +w         &  0.582$\pm$0.009 &    \textbf{0.634$\pm$0.011}      \\
        \hline
    \end{tabular}
\end{minipage}%
  \hfill
  \begin{minipage}[t]{0.48\textwidth}
    \vspace{0pt}
    \centering
        \begin{tabular}{@{}l c c@{}}
        \hline
        \textbf{Method}     & \textbf{Accuracy}  & \textbf{AUROC}  \\
        \hline
        \textit{Supervised Methods}\\
        Sup Base                &   0.713$\pm$0.004        &   0.714$\pm$0.009       \\
        DeepIMV          &   0.737$\pm$0.030      &   0.799$\pm$0.001      \\
        \midrule
        \textit{Unsupervised Methods}\\
        MVAE            &  0.710$\pm$0.000      &   0.577$\pm$0.002      \\
        DMVAE         &   0.710$\pm$0.000         &   0.572$\pm$0.009      \\
        \midrule
        \textit{Semi-supervised Methods}\\
        Semi-sup Base –w      & 0.769$\pm$0.010  & 0.794$\pm$0.007         \\
        CVIB –w   &   0.774$\pm$0.005  &   0.794$\pm$0.007       \\
        MMVM –w   &   \textbf{0.778$\pm$0.001}  &   \textbf{0.805$\pm$0.003}       \\
        Ours –w    &   0.776$\pm$0.006   &   0.793$\pm$0.002       \\
        \hspace{2em} +w  & 0.774$\pm$0.002   &  0.792$\pm$0.005       \\
        \hline
    \end{tabular}
\end{minipage}%
  \hfill
  \begin{minipage}[t]{0.48\textwidth}
    \vspace{0pt}
    \centering
  
\end{minipage}%
\end{table*}

\subsubsection{Cross-modal Generation}
\label{sec:mnist_crossgen}
\vspace{-2ex}

We demonstrate that Cross-view MI regularization offers good reconstruction quality as compared to CVIB regularization, which performs the best among baselines in latent classification.
To boost generation performance and fair comparison, we use the MoPoE framework with the new ELBO derived in Section~\ref{sec:new_elbo_mopoe}, and additionally use a DDPM model~\citep{ho2020denoising} for the prior distribution. 
Furthermore, we observe additional benefit from a generative augmentation regularization from~\citet{zhang2026idmvae} that promotes disentanglement between $z$ and $w_v$ (see detailed discussions in Appendix~\ref{sec:augmi}).

We provide samples of conditional generations in Figure~\ref{fig:tpmnist_gen}.
Without private variable, $z$ learned with PoE has to capture both digit and background during training, and since the digit consists of a small number of pixels of the image, the generations sometimes contain no digit. When private variables and MoPoE are used but the prior is weak (i.e., simple Gaussian), generations have no background. Using MoPoE and diffusion prior together preserves both clear digit identity consistent with the conditions and background, while the counterpart in CVIB cannot preserve digit identity as well in its generations.
This could be because private information is destructive to CVIB's view agreement regularization.
Since we use the same $w$ for each row of generation, we observe that our model achieves disentanglement: images in the same row have the same writing style and location of the digits, which are variations private to each view. 

\subsection{UR-FUNNY and CMU-MOSEI}
\vspace*{-2ex}

We further demonstrate that our method performs competitively against other semi-supervised baselines in two affective computing datasets from MultiBench~\citep{liang2023multizoo}, namely UR-FUNNY~\citep{hasan2019urfunny} and CMU-MOSEI~\citep{zadeh2018multimodal}, which contain three modalities: audio, vision, language.
To process these two datasets, we drop the 50\% of the views before making sure every sample has at least one view. Next, 90\% of the labels are dropped to simulate scenarios with severe missing labels. We perform binary classification of sentiments on both datasets across three random seeds for training. 
We extract private variables from each time step of the time-series data with an RNN, while the shared variable is computed at the sequence level, by averaging another RNN encoder's outputs across time. For likelihood modeling, we broadcast the extracted shared variable to the original time length and concatenate it with the private variable at each time step. The resulting vector is passed through a feedforward network decoder to obtain the per step generation. 
Implementation details can be found in Appendix \ref{apdx:multibench}.

As seen in Table~\ref{tab:pred_multibench}, DeepIMV 
performs better against supervised base and unsupervised methods for both datasets, and our method performs competitively against other semi-supervised baselines. Notably, introducing modality-specific information $w_v$ does not contribute to better classification performance on $z$ for these datasets, as evidenced by unsupervised methods and our methods. 
Nonetheless, shared representations learned from unsupervised data contains predictive information.

\section{Conclusion}
\label{sec:conclusion}
\vspace*{-2ex}

We have proposed a semi-supervised framework that can handle both high quality prediction and generation with missing views and labels in training data.
Our key contributions include the unification of supervised information bottleneck with unsupervised multi-modal VAEs, improved likelihood modeling and generation with private variables, and comparisons against view-agreement regularizations.
In the future, we would like to relax the assumption that predictive information is shared across views, 
and consider scenarios where synergy exists among views~\citep{liang2023pid}. We may also extending the aggregation scheme to be more flexible (e.g., with self-attention,~\citealp{hirt2024perminv}).

\cleardoublepage
\bibliography{multiview}

\newpage

\onecolumn

\title{No Data Wasted: A Semi-supervised Generative Model for Incomplete Multi-view Data Integration with Missing Labels\\(Supplementary Material)}
\maketitle
\appendix
\section{Derivation of ELBO using PoE on a subset of views}
\label{apdx:elbo}
Let $\mathcal{S}$ be the set of available views. We derive a posterior of the shared variable based on a subset of views $\mathcal{O} \subset \mathcal{S}$ using the PoE scheme (cf.~\eqref{eq:poe}):
\begin{equation}
    \begin{aligned}
        q_\theta(z|X_{\mathcal{O}}) 
        \propto p(z) \prod_{v\in \mathcal{O}} q_{\theta}(z|x_v).
    \end{aligned}
\end{equation}
We use $w_v$ to denote the private variable for view $v$, with prior $p(w_v)$ and posterior $q(w_v|x_v)$.

We first derive an ELBO for $\log p(X_{\mathcal{S}})$ using $q(z|X_{\mathcal{O}})$:
\begin{align}\nonumber
    \log p (X_{\mathcal{S}}) & = \log \int p(z,X_{\mathcal{S}}) dz \\ \nonumber
    & = \log \int \frac{p(z,X_{\mathcal{S}})}{q(z|X_{\mathcal{O}})} q(z|X_{\mathcal{O}}) dz \\ \nonumber
    & \ge  \int \log \left[ \frac{p(z,X_{\mathcal{S}})}{q(z|X_{\mathcal{O}})} \right] q(z|X_{\mathcal{O}}) dz \\ 
    \nonumber
    & =  \int \log \left[ \frac{p(X_{\mathcal{S}}|z) p(z)}{q(z|X_{\mathcal{O}})} \right] q(z|X_{\mathcal{O}}) dz \\ 
    \label{eq:elbo-S-1}
    & = \Ex_{q(z|X_{\mathcal{O}})} \left[ \log {p(X_{\mathcal{S}}|z)} \right] - \KL[q(z|X_{\mathcal{O}})||p(z))
\end{align}

where the third step is due to Jensen's inequality.

Next we involve the private variables. Due to the generative assumption that the views are conditionally independent given the shared variable, we have
\begin{align*}
   \log p(X_{\mathcal{S}}|z) = \log \prod_{v\in \mathcal{S}} p(x_v|z) = \sum_{v\in \mathcal{S}} \log \int p(x_v, w_v|z) d w_v.
\end{align*}

To derive lower bound of $\log p(x_v|z)$, we treat the views differently depending on whether it is member of $\mathcal{O}$, the subset used to derive the posterior of $z$.

For $v \in \mathcal{O}$, we have 
\begin{align*}
    \log \int p(x_v, w_v|z) d w_v
    & = \log \int \frac{p(x_v, w_v|z)}{q(w_v|x_v)} q(w_v|x_v) d w_v \\
    & \ge \int \log \left[ \frac{p(x_v, w_v|z)}{q(w_v|x_v)} \right] q(w_v|x_v) d w_v \\
    & = \Ex_{q(w_v|x_v)} \left[ \log {p(x_v|z,w_v)} \right] - \KL[q(w_v|x_v)||p(w_v))
\end{align*}
where in the last step we have used the chain rule $p(x_v, w_v|z) = p(x_v|w_v,z)p(w_v|z)$ and subsequently the assumption that $w_v$ and $z$ are independent (so $p(w_v|z)=p(w_v)$).

For $v \not \in \mathcal{O}$, we have 
\begin{align*}
    \log \int p(x_v, w_v|z) d w_v &= \log \int p(x_v|w_v,z) p(w_v) dw_v \\
    & \ge  \int \log [p(x_v|w_v,z)] p(w_v) dw_v  \\
    & = E_{p(w_v)} [\log  p(x_v|w_v,z)] .
\end{align*}

Combining the above two cases, we continue from~\eqref{eq:elbo-S-1} to obtain
\begin{align}\nonumber
\log p (X_{\mathcal{S}}) \ge 
& \Ex_{z\sim q(z|X_{\mathcal{O}})} \sum_{v\in \mathcal{O}} \left\{ \Ex_{w_v\sim q(w_v|x_v)} [\log p(x_v|z,w_v)]- \KL[q(w_v|x_v)||p(w_v)) \right\} \\
& + \Ex_{z\sim q(z|X_{\mathcal{O}})} \sum_{v \in \mathcal{S} \setminus \mathcal{O}} \left\{ \Ex_{w_v\sim p(w_v)} [\log p(x_v|z,w_v)] \right\} - \KL[q(z|X_{\mathcal{O}})||p(z)] \label{eq:elbo}
\end{align}

In other words, we use posterior sample of $w_v$ in the reconstruction when $v$ is in $\mathcal{O}$, and use prior sample of $w_v$ in the reconstruction otherwise.
The latter case simulates the scenario where only a subset of views are present, and we must use the posterior of $z$ derived from present views along with prior of $w_v$ of the missing view to impute $x_v$ with $p(x_v|z,w_v)$, modeled by the decoder of view $v$.

When we have missing views in the data, the MoPoE scheme uses all subsets of present views for deriving the posterior of $z$, allowing the model to practice present-to-missing imputation. 

\section{Cross-view Mutual Information}\label{apdx:cvmi}

\begin{figure}[b]
    \centering
    \includegraphics[width=0.5\linewidth]{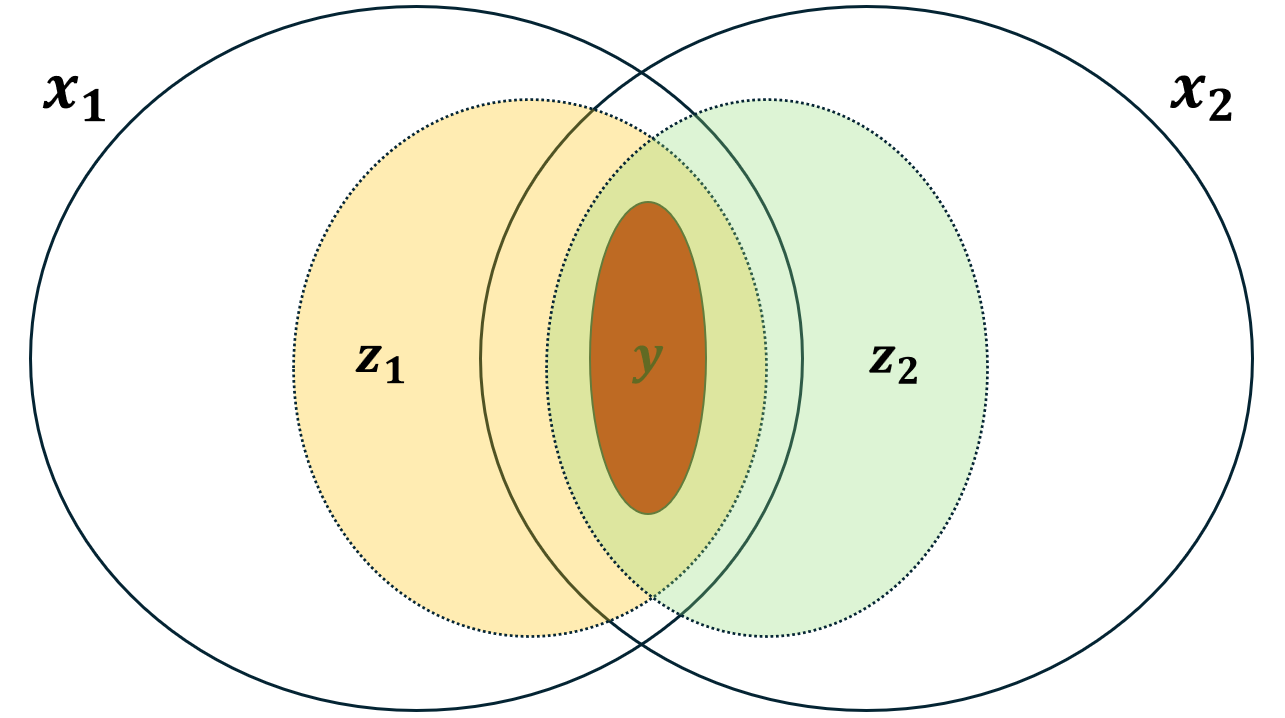}
    \caption{Multi-view redundancy assumption and mutual information decomposition. Since $y$ lies in the intersection of $x_1$ and $x_2$, we encourage $z_1$ to capture information about $x_2$ so as to capture information about $y$.}
    \label{fig:redundancy}
\end{figure}

The unsupervised loss $\cL_{\text{unsup}}$
includes maximization of (a lower bound of) conditional likelihood $p(X|z)$. And minimizing the loss over a training set equivalently minimizes conditional entropy $H(X|z)$, since
\[
\begin{aligned}
H(X|z) &= -\int p(X,z) \log p(X|z) dz dX \\ 
&= \Ex_{X,z} [- \log p(X|z)].
\end{aligned}
\] 
In other words, $\cL_{\text{unsup}}$ encourages $z$ to contain compressed information of $X$, since $I(z; X) = H(X) - H(X|z)$ where $H(X)$ is a constant.
However, pure likelihood modeling leads to potential issues. First, it requires a strong likelihood model $p(X|z)$, which is challenging to obtain for high dimensional structured data, such as high-resolution images.
As a result, pure likelihood modeling may ignore shared information if it only explain a small amount of information for $X$.
While the issue is somewhat alleviated by introducing private variables, having two variables $(z, w_v)$ explain $x_v$ may cause entanglement between them.

Therefore, we introduce additional regularization on $z$ to ensure it captures the shared information across views, from which we predict the label.
In this work, we use a regularization term motivated by the unsupervised multi-view information bottleneck~\citep{federici2020learning} and similar to the one used by~\citet{zhang2026idmvae} with extension to missing view datasets as discussed in Section~\ref{sec:imp_mi}.

To start with, we can decompose the mutual information term between the view and its representation using the chain rule:
\begin{equation}\label{eq:mi}
    I(x_1;z_1) = I(x_1;z_1|y) + I(z_1;y),
\end{equation}
where $z_1$ is the representation of $x_1$ modeled by the posterior $q(z_1|x_1)$,
$I(x_1;z_1|y)$ is the superfluous information (the information in $x_1$ that is not predictive of $y$, ideally captured by $w_1$), and $I(z_1;y)$ is the predictive information. To maximize the expressivity of $z_1$, we can minimize $I(x_1;z_1|y)$ and maximize $I(z_1;y)$ as is the case in the supervised IB framework.

However, when the label $y$ is not observed, one can guarantee that $z_1$ is sufficient for predicting $y$ without knowing $y$, under the multi-view redundancy assumption (Figure \ref{fig:redundancy}).
Therefore, any representation containing all the information shared by both views is as predictive as their joint observation.

This motivates us to use $x_2$ as a proxy of $y$. With this variable substitution, we can instead enforce $x_2$ to contain high mutual information with $z_1$, whose MI is intractable but can be lower bounded:
\begin{equation}\label{eq:mvib_lb}
    \begin{aligned}
        I(z_1;x_2)
        &=I(z_1;z_2 x_2) - I(z_1;z_2|x_2)\\
        &=I(z_1;z_2 x_2)\\
        &=I(z_1;z_2)+I(z_1;x_2|z_2)\\
        &\geq I(z_1;z_2),
    \end{aligned}
\end{equation}
where $I(z_1;z_2|x_2)=0$ due to its variability coming from $x_2$, not $x_1$.

\section{Implementation Details}
All experiments are run on one H100 GPU.
\subsection{TCGA}\label{apdx:tcga}
\subsubsection{Data Processing}
We follow the data processing pipeline described by \citet{lee2021variational}. The 2016 version is processed by combining multi-omics data across different cancer types and different patients before features are selected and kernel PCA was performed to reduce the dimensionality to 100. We obtain a dataset of 10,960 samples (of which 9,477 are labeled) with 5 views. To simulate scenarios with severe missing labels, we first perform 90\%/5\%/5\% training/validation/test splits using five random seeds. On the training set, we drop 50\% views and 95\% and 99\% missing labels, respectively.

Because clinical trials usually have small sample sizes, high dimensionalities, and complex dependencies, this dataset is ideal to test the robustness of our method. One significant outcome of cancer multi-omics data is the days of survival after samples were collected. Because patients may not show up for checkups, it is possible that a sample is censored (unlabeled). The days of survival are then converted to a binary 1-year mortality indicator. The dataset itself has missing views but contains predictive information and high correlations among different views, making it suitable for our task.

\subsubsection{Architecture and Hyperparameters}
We use 2-layer MLP for encoding and decoding, with 128 hidden dimensions, 32 shared dimensions ($z$) and 8 private dimensions ($w$) in the latent space. Learning rate is set to 5e-4, $\alpha=1.0$, and $\gamma=0.01$. $\eta=0.0001, 10.0$ for 5\%,1\% remaining labels, respectively. $\beta=1.0, 0.1$ for 5\%,1\% remaining labels, respectively. $\zeta=1.0, 10.0$ for 5\%,1\% remaining labels, respectively. CVIB loss coefficient is 0.1, and MMVM loss coefficient is 0.1.

\subsection{PolyMNIST-Quadrant}\label{apdx:mnist}

\subsubsection{Data Description}
PolyMNIST-Quadrant \citet{zhang2026idmvae} is an extension of PolyMNIST proposed by \citet{sutter2021generalized}, where MNIST digits are laid on top of patches of randomly cropped five different background. This is made more challenging by randomly putting digit in one of the quadrants. Therefore, a sample has five views, sharing the same digit, which does not necessarily lie in the same quadrant. Our training/validation/test sets have 220,000/5,000/10,000 samples. In the training set, we randomly drop 50\% of views before ensuring at least one view is present per sample. Next, we drop 99.8\% labels for our missing-label setup. We present the performance results on multi-class classification.

\begin{figure}[htbp]
    \centering
    \includegraphics[width=0.5\linewidth]{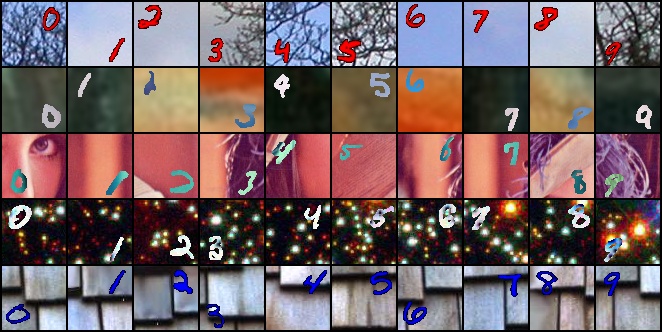}
    \caption{An example of PolyMNIST-Quadrant}
    \label{fig:mnist_example}
\end{figure}

\subsubsection{Architecture and Hyperparameters}
We use the deep residual network (ResNet)~\citep{he2015deepresiduallearningimage} architecture as the encoders and decoders for all methods.
The latent dimension is set to 64 for z and 128 for w.
Learning rate is set to 5e-4. 
$\alpha=200.0$, $\eta=1.0$, $\gamma=1.0$, $\beta=100.0$, $\zeta=200.0$, generative augmentation coefficient is 100.0, CVIB loss coefficient is 2.0. MMVM loss coefficient is 2.0. For generation, the additional diffusion coefficient is 1.0.

\subsubsection{Generative Augmentation and Diffusion}\label{sec:augmi}
Since likelihood model itself is difficult to obtain in high dimensional dataset like this one, for a generation tasks on this dataset we add an generative augmentation scheme proposed in~\citet{zhang2026idmvae} to further disentangle $z$ and $w_v$. This method relies on the intuition that for two samples $X$ and $X'$ in a minibatch, their corresponding latent variables $x_v=(z_v,w_v), x'_v=(z'_v,w'_v)$, and a mapped sample $x_v^+\sim p(\cdot|z_v,w'_v)$, we have similar $q(w_v|x_v^+)\approx q(w_v|x'_v)$. With this intuition, we can define $w_v''\sim q(w_v|x_v^+)$ maximize the contrast between $w'_v$ and $w''_v$.

We use a contrastive loss for matching, i.e.,
\begin{align*}
\cL_{\text{GenAug},w_v} := - Contrast(w_v^{\prime\prime}, w_v^\prime)
\qquad\text{where}\quad 
x_v^+ \sim p(x_v|z_v,w_v^\prime),\;
w_v^{\prime\prime} \sim q(w_v |x_v^+),
\end{align*} 
which approximately minimizes $H (w_v|x_v^\prime)$.
\begin{align*}
\Ex_{x_v\sim p(x_v), x_v^\prime\sim p(x_v), z_v\sim q(z|x_v), w_v^\prime\sim q(w_v|x_v^\prime), x_v^+ \sim p(x_v|z_v,w_v^\prime)}[-\log q(w_v^\prime|x_v^+)].
\end{align*}

We also define $\cL_{\text{GenAug},z_v}$ similarly by switching the role of $z_v$ and $w_v$.
The total redundancy removal regularization is defined as
\begin{align*}
\cL_\text{GenAug}
= \frac{1}{2|\mathcal{S}|}
\sum\nolimits_v \left( \cL_{\text{GenAug},z_v} 
+ \cL_{\text{GenAug},w_v} \right).
\end{align*}

The diffusion loss, proposed in \citet{rombach2021highresolution}, is used to enhance generative quality by performing diffusion reconstruction in latent space on $z$ before decoding it. 
\[
\cL_{\text{LDM}} := \Ex_{z,\epsilon\sim \cN(0,1),t} ||\epsilon-\epsilon_\Omega(z_t,t)||_2^2,
\]
where $\epsilon$ is noise, $t$ is the timestep, and $\Omega$ is the parameters for noise prediction.

\subsection{UR-FUNNY and CMU-MOSEI}\label{apdx:multibench}

\subsubsection{Data Processing}
We use the existing data splits and processing pipeline provided by \citet{liang2023multizoo}, with modifications to drop 50\% of views and 90\% of labels. UR-FUNNY only has binary classification labels on funny or not, and continuous sentiment label, ranging from -3 to 3 of CMU-MOSEI were converted to binary, using 0 as the threshold.

\subsubsection{Architecture and Hyperparameters}
We use GRU (an alternative to RNN) that comes with \citet{liang2023multizoo}'s implementation to encode and a feedforward network to decode time-series data. 

For UR-FUNNY, the latent dimension is set to 32 for z and 8 for w.
Learning rate is 5e-4. $\eta=1.0$, $\alpha=\beta=1.0$, $\gamma=0.01$, and $\zeta=0.08$ for UR-FUNNY. The coefficient for CVIB regularization is 0.001, and the coefficient for MMVM regularization is 0.001. 

For CMU-MOSEI, the latent dimension is set to 32 for both z and w. Learning rate is 5e-4. $\eta=0.0001$, $\alpha=\beta=1.0$, $\gamma=0.01$, and $\zeta=0.01$. The coefficient for CVIB regularization is 0.0001, and the coefficient for MMVM regularization 0.1. 


\section{Imputed Cross-view MI}\label{apdx:ablation}

In Section~\ref{sec:cvmi}, we implemented an imputation heuristic. We define $\ell_{\text{imp}}$ as the loss component of $\cL_{\text{cvmi}}$, and fix its weights for TCGA and UR-FUNNY to be 0.1, while that for CMU-MOSEI to be 10.0.
The ablation study below shows that when we remove the imputed CVMI loss, the performance degrades.

\begin{table}[htbp]
\caption{TCGA with 5\% (left) and 1\% (right) available labels}
    \begin{minipage}[t]{0.48\textwidth}
    \vspace{0pt}
    \centering
    \begin{tabular}{@{}l c c@{}}
    \hline
        \textbf{Method}      & \textbf{Accuracy}  & \textbf{AUROC} \\
        \hline
        Ours $-\ell_{\text{imp}}$  &   0.708$\pm$0.020 &   0.771$\pm$0.022         \\
        Ours  & \textbf{0.717$\pm$0.025} & \textbf{0.779$\pm$0.011} \\
        \hline
    \end{tabular}
    \end{minipage}
    \begin{minipage}[t]{0.48\textwidth}
    \vspace{0pt}
    \centering
    \begin{tabular}{@{}l c c@{}}
    \hline
        \textbf{Method}    & \textbf{Accuracy}    & \textbf{AUROC} \\
        \hline
        Ours $-\ell_{\text{imp}}$   & 0.675$\pm$0.046 & 0.698$\pm$0.028  \\
        Ours  & \textbf{0.677$\pm$0.040} & \textbf{0.718$\pm$0.052} \\
        \hline
    \end{tabular}
    \end{minipage}
\end{table}

\begin{table}[htbp]
\caption{UR-FUNNY and CMU-MOSEI with 10\% available labels.}
    \begin{minipage}[t]{0.48\textwidth}
    \vspace{0pt}
    \centering
    \begin{tabular}{@{}l c c@{}}
    \hline
        \textbf{Method}      & \textbf{Accuracy}  & \textbf{AUROC} \\
        \hline
        Ours $-\ell_{\text{imp}}$      &  0.580$\pm$0.013 &    0.632$\pm$0.010         \\
        Ours       &  \textbf{0.582$\pm$0.009} &    \textbf{0.634$\pm$0.011}        \\
        \hline
    \end{tabular}
    \end{minipage}
    \begin{minipage}[t]{0.48\textwidth}
    \vspace{0pt}
    \centering
    \begin{tabular}{@{}l c c@{}}
    \hline
        \textbf{Method}      & \textbf{Accuracy} & \textbf{AUROC}  \\
        \hline
        Ours $-\ell_{\text{imp}}$     & 0.772$\pm$0.007&  \textbf{0.793$\pm$0.005}       \\
        Ours   & \textbf{0.774$\pm$0.002} &  0.792$\pm$0.005       \\
        \hline
    \end{tabular}
    \end{minipage}
\end{table}
\section{More PolyMNIST-Quadrant Generation}
\begin{figure}[t]
\caption{
PolyMNIST-Quadrant conditional generation (missing view imputation). Conditioned on views 1 (row 1), we generate view 5. Each column correspond to the same sample. Each row share the same private variable.}
\label{fig:tpmnist_gen}
    \centering
\begin{tabular}{c}
Condition, view 1 \\
\includegraphics[width=\linewidth]{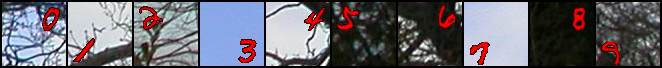} \\
Ours -w, PoE scheme \\
\includegraphics[width=\linewidth]{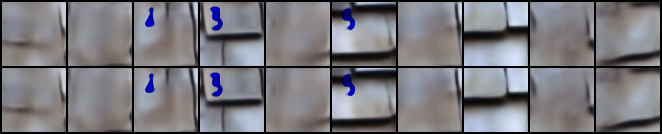} \\
Ours +w, MoPoE scheme, Simple prior for $w_v$ \\
\includegraphics[width=\linewidth]{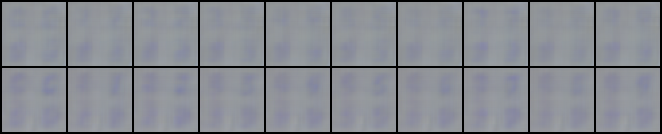} \\
Ours +w, MoPoE scheme, Diffusion prior for $w_v$ \\
\includegraphics[width=\linewidth]{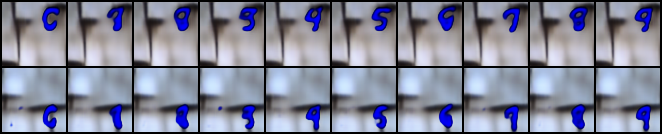} \\
CVIB +w, MoPoE scheme, Diffusion prior for $w_v$ \\
\includegraphics[width=\linewidth]{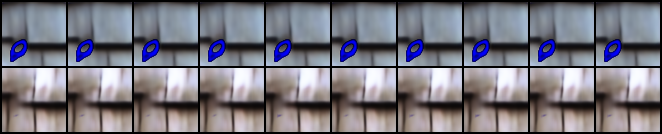}
\end{tabular}
\end{figure}

\begin{figure}[t]
\caption{
PolyMNIST-Quadrant conditional generation (missing view imputation). Conditioned on views 1 and 2 (top 2 rows), we generate view 5. Each column correspond to the same sample. Each row share the same private variable.}
\label{fig:tpmnist_gen}
    \centering
\begin{tabular}{c}
Condition, views 1-2 \\
\includegraphics[width=\linewidth]{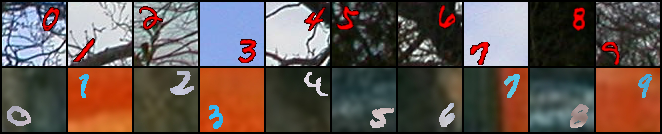} \\
Ours -w, PoE scheme \\
\includegraphics[width=\linewidth]{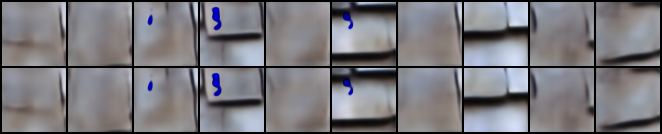} \\
Ours +w, MoPoE scheme, Simple prior for $w_v$ \\
\includegraphics[width=\linewidth]{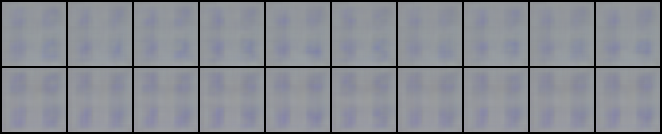} \\
Ours +w, MoPoE scheme, Diffusion prior for $w_v$ \\
\includegraphics[width=\linewidth]{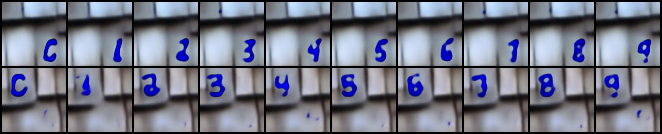} \\
CVIB +w, MoPoE scheme, Diffusion prior for $w_v$ \\
\includegraphics[width=\linewidth]{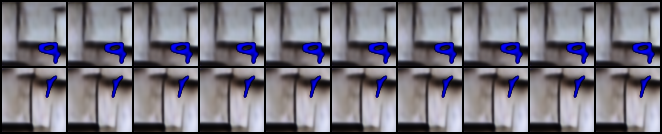}
\end{tabular}
\end{figure}

\begin{figure}[t]
\caption{
PolyMNIST-Quadrant conditional generation (missing view imputation). Conditioned on views 1, 2, and 3 (top 3 rows), we generate view 5. Each column correspond to the same sample. Each row share the same private variable.}
\label{fig:tpmnist_gen}
    \centering
\begin{tabular}{c}
Condition, views 1-3 \\
\includegraphics[width=\linewidth]{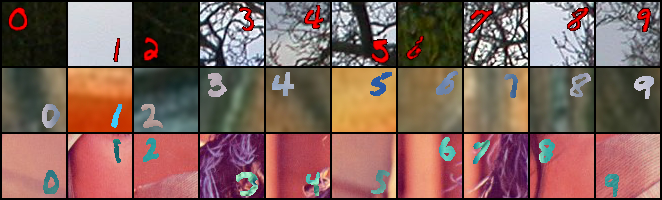} \\
Ours -w, PoE scheme \\
\includegraphics[width=\linewidth]{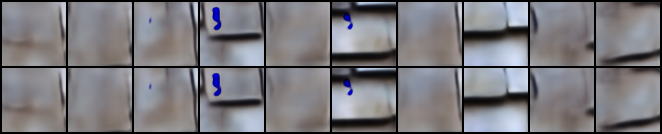} \\
Ours +w, MoPoE scheme, Simple prior for $w_v$ \\
\includegraphics[width=\linewidth]{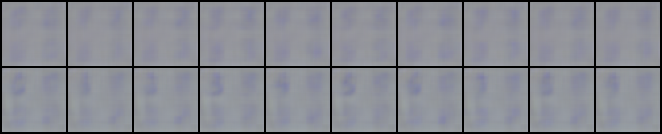} \\
Ours +w, MoPoE scheme, Diffusion prior for $w_v$ \\
\includegraphics[width=\linewidth]{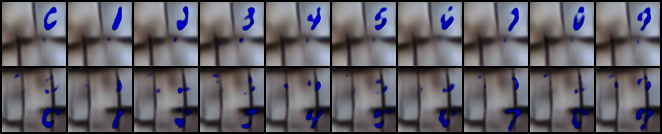} \\
CVIB +w, MoPoE scheme, Diffusion prior for $w_v$ \\
\includegraphics[width=\linewidth]{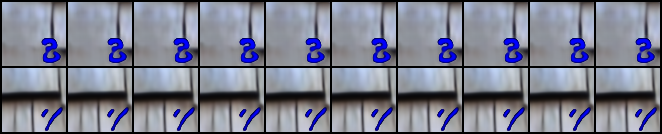}
\end{tabular}
\end{figure}

\end{document}